\documentclass[11pt]{article}

\usepackage[margin=1in]{geometry}

\usepackage{iftex}

\ifPDFTeX
  \usepackage[T1]{fontenc}
  \usepackage[utf8]{inputenc} 
  \usepackage{lmodern}

\else
  \usepackage{fontspec}
  \newfontfamily\EmojiFont{DejaVu Sans}
  \IfFontExistsTF{Noto Color Emoji}{%
    \renewfontfamily\EmojiFont{Noto Color Emoji}[Renderer=HarfBuzz]%
  }{%
    \IfFontExistsTF{Symbola}{\renewfontfamily\EmojiFont{Symbola}}{}%
  }

\fi

\usepackage{amsmath,amssymb,bm}

\usepackage{graphicx}

\usepackage{booktabs}
\usepackage{multirow}
\usepackage{array}
\usepackage{colortbl}

\usepackage[dvipsnames,table]{xcolor}

\definecolor{bestbg}{HTML}{D4EDDA}      
\definecolor{secondbg}{HTML}{FFF3CD}    
\definecolor{besttext}{HTML}{155724}    
\definecolor{secondtext}{HTML}{856404}  

\definecolor{samgreen}{HTML}{059669}

\usepackage{enumitem}

\usepackage[
  colorlinks=true,
  linkcolor=MidnightBlue,
  citecolor=MidnightBlue,
  urlcolor=MidnightBlue
]{hyperref}

\usepackage[font=small,labelfont=bf]{caption}

\usepackage{fancyhdr}
\newcommand{\best}[1]{\cellcolor{bestbg}\textbf{\textcolor{besttext}{#1}}}
\newcommand{\second}[1]{\cellcolor{secondbg}\textcolor{secondtext}{#1}}
\newcommand{\modelname}{\textsc{GenSeg-R1}}

\title{%
  \vspace{-1em}
  \textbf{GenSeg-R1: RL-Driven Vision--Language Grounding\\[4pt]
  for Fine-Grained Referring Segmentation}%
  \vspace{-0.5em}
}
\author{%
  \begin{tabular}{cccc}
    Sandesh Hegde\thanks{Equal contribution.}
    & Jaison Saji Chacko\footnotemark[1]
    & Debarshi Banerjee
    & Uma Mahesh \\[2pt]
    {\ttfamily\small sandesh.hegde@camcom.ai}
    & {\ttfamily\small jaison.saji@camcom.ai}
    & {\ttfamily\small debarshi.banerjee@camcom.ai}
    & {\ttfamily\small umesh@camcom.ai} \\[4pt]
    \multicolumn{4}{c}{\small Camcom Technologies Pvt.\ Ltd.}
  \end{tabular}%
}
\date{}

\begin{document}
\maketitle
\thispagestyle{fancy}

\begin{abstract}
\noindent
We study fine-grained referring image segmentation via a decoupled
\emph{reason-then-segment} pipeline.  A vision--language model (VLM)
receives an image and a natural-language query, reasons about the
scene, and emits structured spatial prompts---a bounding box plus two
interior keypoints---for every referred instance.  A frozen
promptable segmenter (SAM\,2) converts these prompts into
high-quality masks.

Within our \modelname{} framework we finetune Qwen3-VL models (4\,B
and 8\,B parameters) using Group Relative Policy Optimization (GRPO),
requiring \emph{no} supervised reasoning-chain annotations.  On
RefCOCOg validation our best model (\modelname{}-8B) achieves
\textbf{0.7127} cIoU and \textbf{0.7382} mIoU, substantially
outperforming the corresponding Qwen3-VL Instruct baselines
(\,+15.3 and +21.9 points, respectively) and surpassing
Seg-Zero-7B~\cite{segzero} by +3.3 cIoU under identical evaluation.

We further introduce \modelname{}-G, a variant trained on
GRefCOCO~\cite{grefcoco} with a SAM\,2-in-the-loop reward that
directly optimizes mask quality.  On GRefCOCO validation
\modelname{}-G achieves \textbf{76.69\%} target mIoU with
\textbf{82.40\%} accuracy on negative (no-target) prompts,
substantially outperforming Seg-R1-7B and Seg-Zero-7B which entirely
lack no-target detection capability.  On ReasonSeg test,
\modelname{}-4B reaches \textbf{68.40\%} mIoU, surpassing
Seg-Zero-7B by +7.0 and Seg-R1-7B by +10.7 points.
\end{abstract}

\section{Introduction}\label{sec:intro}

Fine-grained segmentation from natural language is a core capability
for interactive vision systems, robotics, and assistive tools.  A
predominant paradigm is to first \emph{ground} a textual query in the
image---typically as a bounding box, point(s), or region
proposal---and then pass this spatial prompt to a strong promptable
segmentation model~\cite{sam2} to obtain a precise mask.

Despite rapid progress, many existing grounding models exhibit a
critical reliability issue: they often fail to respect
\emph{negative} queries.  For example, given an image of a person
wearing a red shirt and the prompt \emph{``person with blue shirt,''}
prior models may still localize the visible person even though the
query is inconsistent with the image.  In downstream segmentation
this produces confident but \emph{incorrect} masks---a worse outcome
than returning nothing.

This paper presents \modelname{}, an improved grounding model
tailored to fine-grained segmentation pipelines.  Our approach
combines three ingredients: (i)~stronger VLM backbones (Qwen3-VL),
(ii)~GRPO training that directly optimizes grounding actions via a
SAM\,2-in-the-loop reward, and (iii)~explicit handling of negative
and empty-target prompts through GRefCOCO~\cite{grefcoco}.

\paragraph{Contributions.}
\begin{itemize}[leftmargin=*,itemsep=3pt]
  \item An improved grounding model based on Qwen3-VL, replacing
        Qwen2.5-VL used in prior recipes, yielding stronger spatial
        reasoning out of the box.
  \item A GRPO training procedure whose reward is computed by running
        SAM\,2 on the model's predicted box\,+\,keypoints, measuring
        mask IoU against ground truth.  This closes the loop between
        the VLM's textual output and the final segmentation quality.
  \item \modelname{}-G: a variant trained on GRefCOCO with explicit
        negative-prompt handling and a SAM\,2-in-the-loop reward,
        achieving 82.40\% no-target accuracy with only 0.5\%
        false-negative rate.
  \item State-of-the-art results on RefCOCOg validation under
        identical evaluation protocol, outperforming Seg-Zero-7B
        (+3.3 cIoU) and Seg-R1-7B (+5.2 cIoU), with consistent gains
        on GRefCOCO validation and the reasoning-heavy ReasonSeg
        benchmark.
\end{itemize}

\section{Related Work}\label{sec:related}

\paragraph{Promptable segmentation.}
The Segment Anything Model (SAM)~\cite{sam2} and its successor SAM\,2
establish the paradigm of high-quality mask prediction conditioned on
spatial prompts such as boxes, points, and coarse masks.  Because
SAM-family models are trained on massive data and generalize across
domains, they serve as an effective \emph{frozen} segmentation
back-end that can be paired with upstream localization modules.

\paragraph{Grounding for segmentation.}
Several recent works translate textual queries into spatial prompts
consumable by SAM-like models.  Seg-Zero~\cite{segzero} uses
GRPO-trained VLM prompting with Qwen2.5-VL-7B; Seg-R1~\cite{segr1}
and SAM-R1~\cite{samr1} explore reinforcement-style training recipes
for improved grounding.  Our \modelname{} framework shares this
\emph{reason-then-segment} philosophy but introduces a SAM\,2-in-the-loop
reward and explicit negative-prompt handling.

\paragraph{Referring expression benchmarks.}
RefCOCO, RefCOCO+, and RefCOCOg~\cite{refcoco} are classic
single-object referring expression datasets.
GRefCOCO~\cite{grefcoco} extends the setting with multi-target
references and negative prompts with empty targets, making it better
suited for evaluating robustness to non-matching queries.

\paragraph{Instruction data for grounding.}
LISA~\cite{lisa} provides instruction-following data that aligns
language with segmentation-relevant outputs.  Our data recipe draws on
VisionReasoner multi-object annotations combined with GRefCOCO
negative examples.

\section{Method}\label{sec:method}

\subsection{Pipeline Overview}

\begin{figure}[t]
\centering
\includegraphics[width=\textwidth]{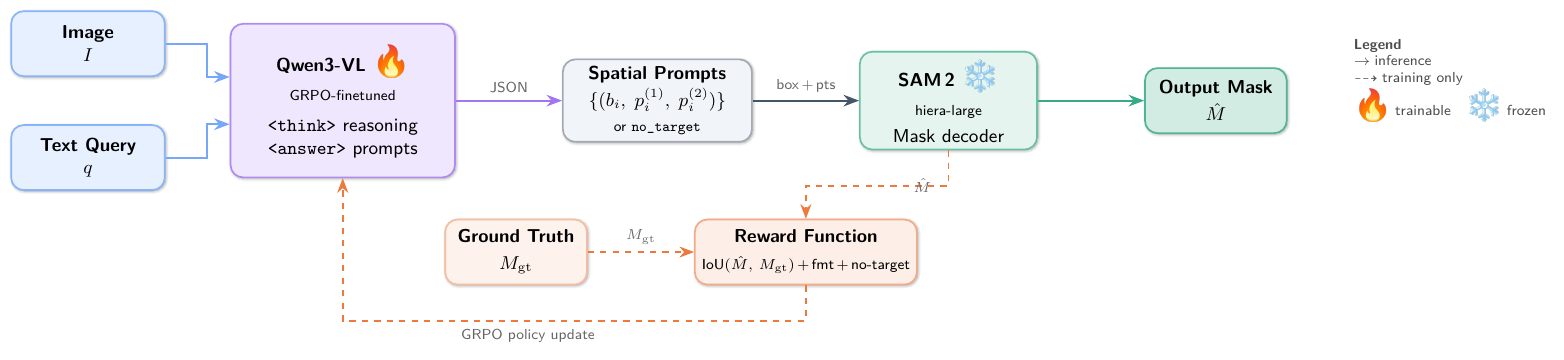}
\caption{%
  \textbf{\modelname{} architecture.}
  \emph{Stage~1:} Qwen3-VL (trainable) processes the image
  and query, reasons in \texttt{<think>} tags, and outputs structured
  spatial prompts (bounding box + two keypoints) or a
  \texttt{no\_target} flag.
  \emph{Stage~2:} SAM\,2 (frozen) converts prompts into
  masks.  During GRPO training (dashed arrows),
  the SAM\,2 mask IoU feeds back as a reward signal to update the VLM
  policy.}
\label{fig:arch}
\end{figure}

Figure~\ref{fig:arch} illustrates the full \modelname{} pipeline.
Given an input image $I$ and a text query $q$, the system operates in
two stages:

\begin{enumerate}[leftmargin=*,label=\textbf{Stage~\arabic*.},itemsep=4pt]
  \item \textbf{Reasoning \& Grounding.}\;
    A finetuned Qwen3-VL model processes $(I, q)$ and emits structured
    spatial prompts inside \texttt{<answer>} tags, optionally preceded
    by free-form reasoning in \texttt{<think>} tags.
  \item \textbf{Promptable Segmentation.}\;
    A frozen SAM\,2 model consumes the predicted box and keypoints for
    each instance and outputs a binary mask.
\end{enumerate}

\noindent
Crucially, no supervised chain-of-thought annotations are required:
reasoning emerges naturally through GRPO training, and we observe
coherent test-time \texttt{<think>} traces in the majority of outputs.

\subsection{Base Models and Coordinate System}

We finetune Qwen3-VL Instruct models at two scales: \textbf{4\,B}
and \textbf{8\,B} parameters.  Qwen3-VL natively supports a
normalized $[0,1000]$ coordinate system; we rescale all training
annotations accordingly (e.g., an $840\times840$ image maps linearly
to $[0,1000]^2$).

\subsection{Output Format: Box + Two Keypoints}

For each predicted instance the model emits a bounding box and two
interior keypoints:
\begin{equation}\label{eq:output}
  \mathcal{P}
  = \bigl\{(b_i,\; p_i^{(1)},\; p_i^{(2)})\bigr\}_{i=1}^{N},
\end{equation}
where $b_i = [x_1, y_1, x_2, y_2]$ and
$p_i^{(1)}, p_i^{(2)} \in [0,1000]^2$.  Two interior points provide
SAM\,2 with richer localization cues than a single centroid, reducing
ambiguity for elongated or multi-part objects.

For negative (no-target) queries the model instead outputs
\texttt{\{"no\_target": true\}}, signalling that no mask should be
produced.

\subsection{SAM\,2 Integration}

We use a frozen SAM\,2 checkpoint
(\texttt{facebook/sam2\text{-}hiera\text{-}large}).  For each instance
$i$ we invoke SAM\,2 with the predicted box $b_i$ as a box prompt and
$p_i^{(1)}$ as a positive-point prompt (optionally both points),
then select the highest-confidence mask.  For multi-object queries we
take the pixel-wise union of instance masks.

\subsection{Reinforcement Learning with GRPO}\label{sec:grpo}

We train the VLM with Group Relative Policy Optimization
(GRPO)~\cite{r1recipe}.  For each training example $(I, q)$ we
sample $n{=}8$ candidate responses from the current policy, score
each with a composite reward (Section~\ref{sec:reward}), and update
the policy toward higher-reward samples.  GRPO avoids the cost of
training a separate value network by normalizing rewards within each
group, making it practical even on moderate hardware.

\subsection{Reward Design}\label{sec:reward}

We use two reward configurations corresponding to the two model
variants.

\paragraph{\modelname{}-4B/8B reward (distance-based).}
A lightweight reward combining: (i)~format validation of
\texttt{<think>} and \texttt{<answer>} tags with valid JSON,
(ii)~bounding-box accuracy (IoU $> 0.5$ and L1 distance thresholds),
(iii)~point accuracy (minimum distance $< 119$ in $[0,1000]$ space,
point inside predicted box), and (iv)~a non-repetition penalty.  This
reward requires no model inference, enabling fast training.

\paragraph{\modelname{}-G reward (SAM\,2-in-the-loop).}
An outcome-based reward that extends the distance-based reward with
three additional components:

\begin{enumerate}[leftmargin=*,label=\textbf{R\arabic*.},itemsep=3pt]
  \item \textbf{SAM\,2 IoU reward}\;($r_{\mathrm{iou}}$, weight 5.0):\;
    We generate a mask with SAM\,2 using the predicted box and points,
    then compute $\mathrm{IoU}(\hat{M}, M_{\mathrm{gt}})$.  This is
    the primary signal that directly ties the VLM output to
    segmentation quality.
  \item \textbf{Negative-point validity}\;($r_{\mathrm{neg}}$, weight 10.0):\;
    Validates that predicted negative points lie inside the bounding box
    but outside the ground-truth mask, at sufficient distance from the
    mask boundary.  This is the highest-weighted component, driving
    boundary refinement.
  \item \textbf{No-target reward}\;($r_{\mathrm{nt}}$, reward 10.0):\;
    For negative queries with empty ground-truth masks, the model is
    rewarded for predicting \texttt{no\_target} and penalized for
    hallucinating boxes.
\end{enumerate}

The SAM\,2-in-the-loop reward is more expensive (SAM\,2 inference per
rollout sample) but directly optimizes for final mask quality rather
than proxy coordinate metrics.

\subsection{Training Data}\label{sec:data}

\paragraph{\modelname{}-4B/8B.}
Trained on \textbf{VisionReasoner-MultiObjects-7K} (7\,099 samples
with bounding-box and single-point annotations).  Images are resized
to $840\times840$ with coordinate rescaling to $[0,1000]$.

\paragraph{\modelname{}-G.}
Trained on \textbf{GRefCOCO with negative points} (15\,285 samples),
which includes both positive referring expressions and negative
(no-target) queries.  Each sample provides two positive interior
points, two negative (background) points, and ground-truth polygon
masks for computing the SAM\,2-in-the-loop reward.  The roughly 2$\times$
larger dataset and richer annotations enable the model to learn both
precise boundary prompting and reliable no-target rejection.

\subsection{Implementation Details}

\begin{itemize}[leftmargin=*,itemsep=2pt]
  \item \textbf{Position encoding:}\;
    Qwen3-VL uses mRoPE; we handle position continuity via
    model-provided RoPE deltas when computing position IDs.
  \item \textbf{Parallelism:}\;
    Distributed training with FSDP (ZeRO-3 sharding).
  \item \textbf{Rollout:}\;
    vLLM for efficient batched sampling of $n{=}8$ completions per
    prompt.
  \item \textbf{Hyperparameters:}\;
    See Table~\ref{tab:hparams}.
\end{itemize}

\begin{table}[t]
  \centering
  \caption{Training hyperparameters.}
  \label{tab:hparams}
  \small
  \begin{tabular}{ll}
    \toprule
    \textbf{Parameter} & \textbf{Value} \\
    \midrule
    Learning rate        & $1\times 10^{-6}$  \\
    Weight decay         & $1\times 10^{-2}$  \\
    Max gradient norm    & 1.0                 \\
    KL coefficient       & $5\times 10^{-3}$ (low-variance KL) \\
    Global batch size    & 8 (micro-batch 2/device) \\
    Rollout batch size   & 8 \\
    Samples per prompt ($n$) & 8 \\
    Max prompt length    & 2\,048 tokens \\
    Max response length  & 1\,500 tokens \\
    \midrule
    \multicolumn{2}{l}{\textit{Hardware}} \\
    \modelname{}-8B      & $4\times$ H200 (141\,GB) \\
    \modelname{}-4B / -G & $2\times$ H200 (141\,GB) \\
    \bottomrule
  \end{tabular}
\end{table}

\section{Experiments}\label{sec:exp}

\subsection{Setup}

\paragraph{Benchmarks.}
We evaluate on three benchmarks:
\textbf{RefCOCOg validation} (2\,573 referring expressions over
1\,300 images),
\textbf{GRefCOCO validation}~\cite{grefcoco} (1\,000 samples: 642 with targets, 358
with no target), and
\textbf{ReasonSeg test}~\cite{lisa} (779 samples requiring compositional
reasoning).
All benchmarks use the same evaluation protocol: the VLM predicts
spatial prompts, SAM\,2 generates masks, and metrics are computed
against ground-truth masks.

\paragraph{Models.}
We compare \modelname{}-8B, \modelname{}-4B, and \modelname{}-G
against un-finetuned Instruct baselines (Qwen3-VL~4B/8B and
Qwen2.5-VL-7B) and recent RL-based segmentation models:
Seg-Zero-7B~\cite{segzero} (GRPO on Qwen2.5-VL-7B),
VisionReasoner-7B (SFT on Qwen2.5-VL-7B), and
Seg-R1-7B/3B~\cite{segr1} (GRPO on Qwen2.5-VL, trained on
DIS5K+COD10K).  All models are evaluated under the \emph{same}
protocol with the same SAM\,2 checkpoint and evaluation code.

\subsection{Metrics}

\paragraph{Cumulative IoU (cIoU).}
Defined as $\mathrm{cIoU} = \sum_i |P_i \cap G_i| \;/\;
\sum_i |P_i \cup G_i|$, which weights each sample by object area.

\paragraph{Mean IoU (mIoU).}
The unweighted average of per-sample IoU values.

\paragraph{Precision at threshold.}
$\mathrm{P}@\tau$ is the fraction of samples with
$\mathrm{IoU} \ge \tau$.  We report $\mathrm{P}@0.5$,
$\mathrm{P}@0.7$, and $\mathrm{P}@0.9$.

\paragraph{Detection metrics.}
For bounding boxes we additionally report AP, AP$@0.5$, and AP$@0.75$.

\paragraph{No-Target Accuracy.}
For GRefCOCO, which includes queries with no matching object, we
report the fraction of no-target samples for which the model correctly
predicts \texttt{no\_target} instead of hallucinating a bounding box.

\subsection{RefCOCOg Validation Results}

Tables~\ref{tab:refcocog_mask} and~\ref{tab:refcocog_box} present
mask and box results, respectively.  All RL-based models are evaluated
under the \emph{same} protocol: same RefCOCOg val split (2\,573
samples), same SAM\,2 (sam2-hiera-large), same evaluation code.
Color coding:
\colorbox{bestbg}{\textcolor{besttext}{\textbf{best}}} and
\colorbox{secondbg}{\textcolor{secondtext}{second best}}.

\begin{table}[t]
  \centering
  \caption{RefCOCOg validation --- \textbf{segmentation} (SAM\,2 mask metrics).
    All RL-based models evaluated under identical protocol.}
  \label{tab:refcocog_mask}
  \small
  \setlength{\tabcolsep}{6pt}
  \begin{tabular}{l l ccccc}
    \toprule
    \textbf{Model} & \textbf{Base VLM}
      & \textbf{cIoU}~$\uparrow$ & \textbf{mIoU}~$\uparrow$
      & \textbf{P@0.5}~$\uparrow$ & \textbf{P@0.7}~$\uparrow$
      & \textbf{P@0.9}~$\uparrow$ \\
    \midrule
    \modelname{}-8B (ours)  & Qwen3-VL-8B
      & \best{0.7127}  & \best{0.7382}  & \best{0.8232}  & \best{0.7314}  & \best{0.3545} \\
    \modelname{}-4B (ours)  & Qwen3-VL-4B
      & \second{0.7062} & \second{0.7318} & \second{0.8166} & \second{0.7120} & \second{0.3533} \\
    \midrule
    Seg-Zero-7B~\cite{segzero} & Qwen2.5-VL-7B
      & 0.6794 & 0.7155 & 0.7913 & 0.6852 & 0.3393 \\
    Seg-R1-7B~\cite{segr1}  & Qwen2.5-VL-7B
      & 0.6607 & 0.7014 & 0.7824 & 0.6774 & 0.3199 \\
    VisionReasoner-7B       & Qwen2.5-VL-7B
      & 0.6530 & 0.6966 & 0.7680 & 0.6533 & 0.3160 \\
    Seg-R1-3B~\cite{segr1}  & Qwen2.5-VL-3B
      & 0.6294 & 0.6724 & 0.7470 & 0.6541 & 0.3129 \\
    \midrule
    Qwen3-VL-8B-Instruct   & ---
      & 0.5596 & 0.5188 & 0.5834 & 0.5192 & 0.2604 \\
    Qwen2.5-VL-7B-Instruct & ---
      & 0.5099 & 0.5276 & 0.5841 & 0.5115 & 0.2534 \\
    Qwen3-VL-4B-Instruct   & ---
      & 0.5019 & 0.4890 & 0.5453 & 0.4885 & 0.2336 \\
    \bottomrule
  \end{tabular}
\end{table}

\begin{table}[t]
  \centering
  \caption{RefCOCOg validation --- \textbf{bounding box} detection.}
  \label{tab:refcocog_box}
  \small
  \setlength{\tabcolsep}{8pt}
  \begin{tabular}{l cccc}
    \toprule
    \textbf{Model} & \textbf{bbox mIoU}~$\uparrow$ & \textbf{AP}~$\uparrow$
      & \textbf{AP@0.5}~$\uparrow$ & \textbf{AP@0.75}~$\uparrow$ \\
    \midrule
    \modelname{}-8B  (ours)
      & \second{0.8028} & \second{0.7277} & \second{0.8717} & \second{0.7835} \\
    \modelname{}-4B  (ours)
      & \best{0.8101}  & \best{0.7347}  & \best{0.8780}  & \best{0.7890} \\
    Seg-Zero-7B~\cite{segzero}
      & 0.8000 & 0.7197 & --- & --- \\
    VisionReasoner-7B
      & 0.7967 & 0.7149 & 0.8581 & 0.7602 \\
    Seg-R1-7B~\cite{segr1}
      & 0.7452 & 0.6294 & 0.8216 & 0.7015 \\
    \midrule
    Qwen2.5-VL-7B-Instruct
      & 0.5826 & 0.5162 & 0.6273 & 0.5515 \\
    Qwen3-VL-8B-Instruct
      & 0.5546 & 0.4981 & 0.5993 & 0.5325 \\
    Qwen3-VL-4B-Instruct
      & 0.5301 & 0.4745 & 0.5702 & 0.5049 \\
    \bottomrule
  \end{tabular}
\end{table}

\paragraph{Key observations.}
\modelname{}-8B improves over the Qwen3-VL-8B Instruct baseline by
\textbf{+15.3} cIoU and \textbf{+21.9} mIoU points on RefCOCOg
validation masks.  The 4B variant achieves comparable improvements
(\textbf{+20.4} cIoU, \textbf{+24.3} mIoU) over its smaller
baseline, demonstrating that GRPO is effective even at moderate model
scale.

Under identical evaluation, \modelname{}-8B outperforms
Seg-Zero-7B~\cite{segzero} by \textbf{+3.3} cIoU and \textbf{+2.3}
mIoU, and Seg-R1-7B~\cite{segr1} by \textbf{+5.2} cIoU and
\textbf{+3.7} mIoU.  Notably, \modelname{}-4B (4\,B parameters)
outperforms Seg-Zero-7B (7\,B) by +2.7 cIoU despite being nearly
half the size, suggesting that the Qwen3-VL backbone provides
stronger spatial reasoning capabilities than Qwen2.5-VL.

On bounding-box metrics the 4B model slightly exceeds the
8B model (0.8101 vs.\ 0.8028 bbox mIoU).  We attribute this to the
8B model occasionally predicting tighter boxes that sacrifice box IoU
for better interior-point placement, ultimately yielding superior mask
quality.

\subsection{SAM\,2 Prompt Ablation}\label{sec:ablation_sam}

Table~\ref{tab:sam_ablation} isolates the contribution of interior
keypoints by varying the SAM\,2 prompt mode at inference time while
keeping the same \modelname{}-4B model.

\begin{table}[t]
  \centering
  \caption{SAM\,2 prompt ablation on RefCOCOg validation using
    \modelname{}-4B.  Interior keypoints provide monotonic
    improvement, especially at the strictest threshold.}
  \label{tab:sam_ablation}
  \small
  \setlength{\tabcolsep}{8pt}
  \begin{tabular}{l ccccc}
    \toprule
    \textbf{SAM\,2 Prompt Mode}
      & \textbf{cIoU}~$\uparrow$ & \textbf{mIoU}~$\uparrow$
      & \textbf{P@0.5}~$\uparrow$ & \textbf{P@0.7}~$\uparrow$
      & \textbf{P@0.9}~$\uparrow$ \\
    \midrule
    Box only
      & 0.6929 & 0.7223 & 0.8072 & 0.7035 & 0.3385 \\
    Box + 1 interior point
      & 0.7013 & 0.7268 & 0.8138 & 0.7108 & 0.3498 \\
    \textbf{Box + 2 interior points}
      & \best{0.7062} & \best{0.7318} & \best{0.8166} & \best{0.7120} & \best{0.3533} \\
    \bottomrule
  \end{tabular}
\end{table}

Interior keypoints provide consistent, monotonic improvements across
all metrics.  The gain is most pronounced at the strictest threshold
(P@0.9: +0.0148 from box-only to box+2pt), indicating that
keypoints primarily improve fine-grained mask quality.  Even a single
interior point provides meaningful improvement (+0.0084 cIoU), with
the second point contributing an additional +0.0049 cIoU.

\subsection{GRefCOCO: Negative Prompt Results}\label{sec:grefcoco}

Table~\ref{tab:grefcoco} evaluates models on GRefCOCO
validation, which includes both positive (target-present) and negative
(no-target) queries.

\begin{table}[t]
  \centering
  \caption{GRefCOCO validation results (1\,000 samples: 642 target,
    358 no-target).  \modelname{}-G is trained on GRefCOCO with the
    SAM\,2-in-the-loop reward.  Seg-R1 and Seg-Zero lack no-target
    training and hallucinate masks on all negative queries.}
  \label{tab:grefcoco}
  \small
  \setlength{\tabcolsep}{5pt}
  \begin{tabular}{l cc c}
    \toprule
    \textbf{Model}
      & \textbf{Target mIoU (\%)}~$\uparrow$
      & \textbf{Target cIoU (\%)}~$\uparrow$
      & \textbf{No-Target Acc (\%)}~$\uparrow$ \\
    \midrule
    \modelname{}-G\; (ours)
      & \best{76.69}   & \best{76.83}   & \best{82.40} \\
    \modelname{}-4B (ours)
      & \second{75.58}  & \second{75.54}  & \second{79.05} \\
    \midrule
    Seg-R1-7B~\cite{segr1}
      & 69.51  & 69.02  & 0.00 \\
    Seg-Zero-7B~\cite{segzero}
      & 66.11  & 65.90  & 0.00 \\
    \bottomrule
  \end{tabular}
\end{table}

\paragraph{Key findings.}
\modelname{}-G achieves the highest target mIoU (76.69\%) and
no-target accuracy (82.40\%), surpassing Seg-R1-7B by \textbf{+7.2}
mIoU points and Seg-Zero-7B by \textbf{+10.6} points on target
queries.  More strikingly, Seg-R1 and Seg-Zero achieve \textbf{0\%}
no-target accuracy---they hallucinate masks for every negative
query---while \modelname{}-G correctly abstains on 295/358
no-target prompts.  \modelname{}-4B, trained on VisionReasoner data
without explicit GRefCOCO no-target examples, still achieves 79.05\%
no-target accuracy, indicating that the GRPO format reward alone
provides some no-target capability.

The false-negative rate (predicting no-target when the object exists)
is very low: only 3/642 (0.5\%) for \modelname{}-G and 12/642 (1.9\%)
for \modelname{}-4B, compared to 0/642 for Seg-R1 and Seg-Zero (which
never predict no-target).

\begin{figure*}[t]
  \centering
  \includegraphics[width=\textwidth]{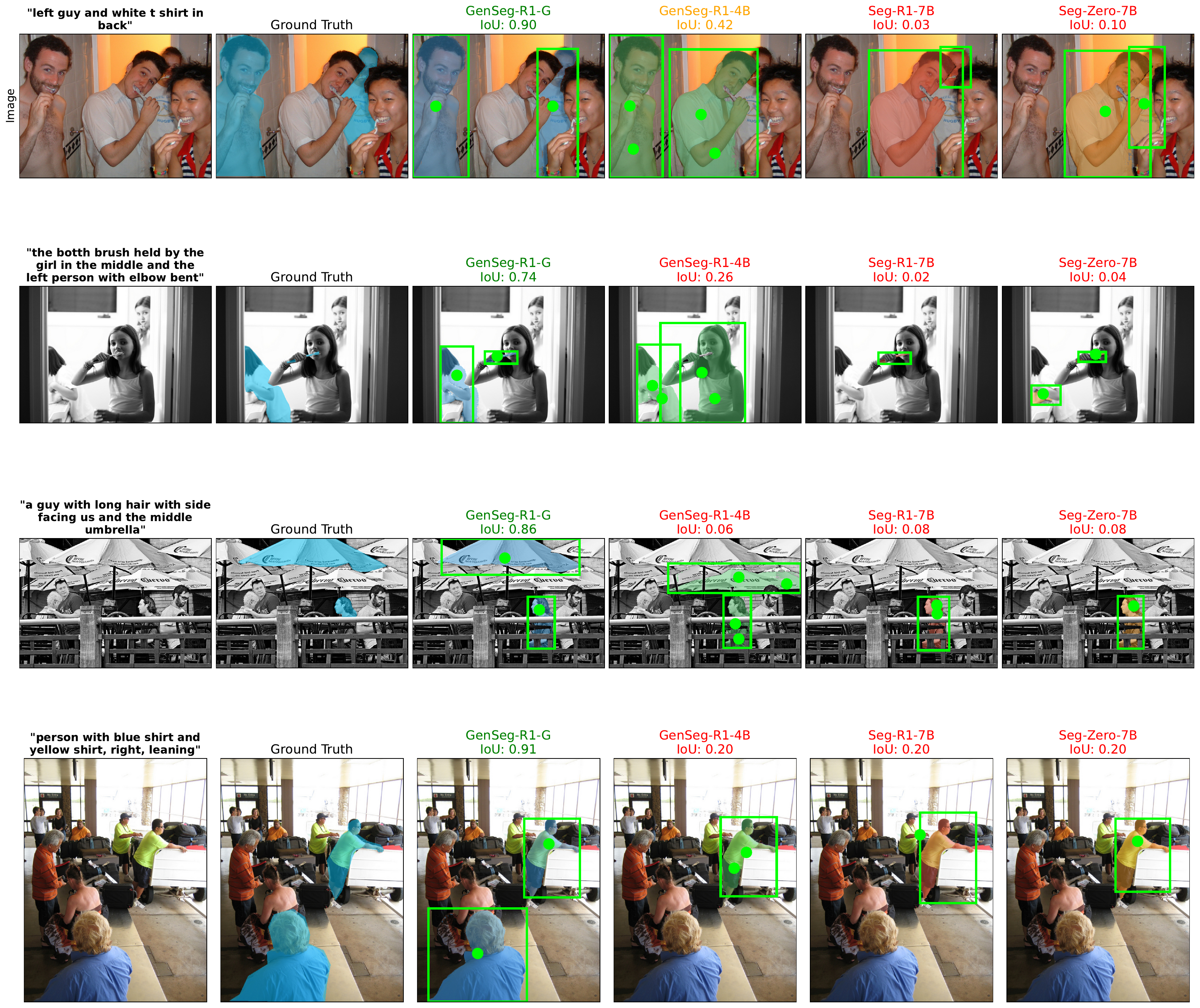}
  \caption{\textbf{Qualitative segmentation comparison.}
    Each row shows a different query.  \modelname{}-G and \modelname{}-4B
    produce accurate masks (green overlay) closely matching ground truth,
    while Seg-R1-7B and Seg-Zero-7B produce poor or misaligned masks.}
  \label{fig:qual_seg}
\end{figure*}

\begin{figure*}[t]
  \centering
  \includegraphics[width=\textwidth]{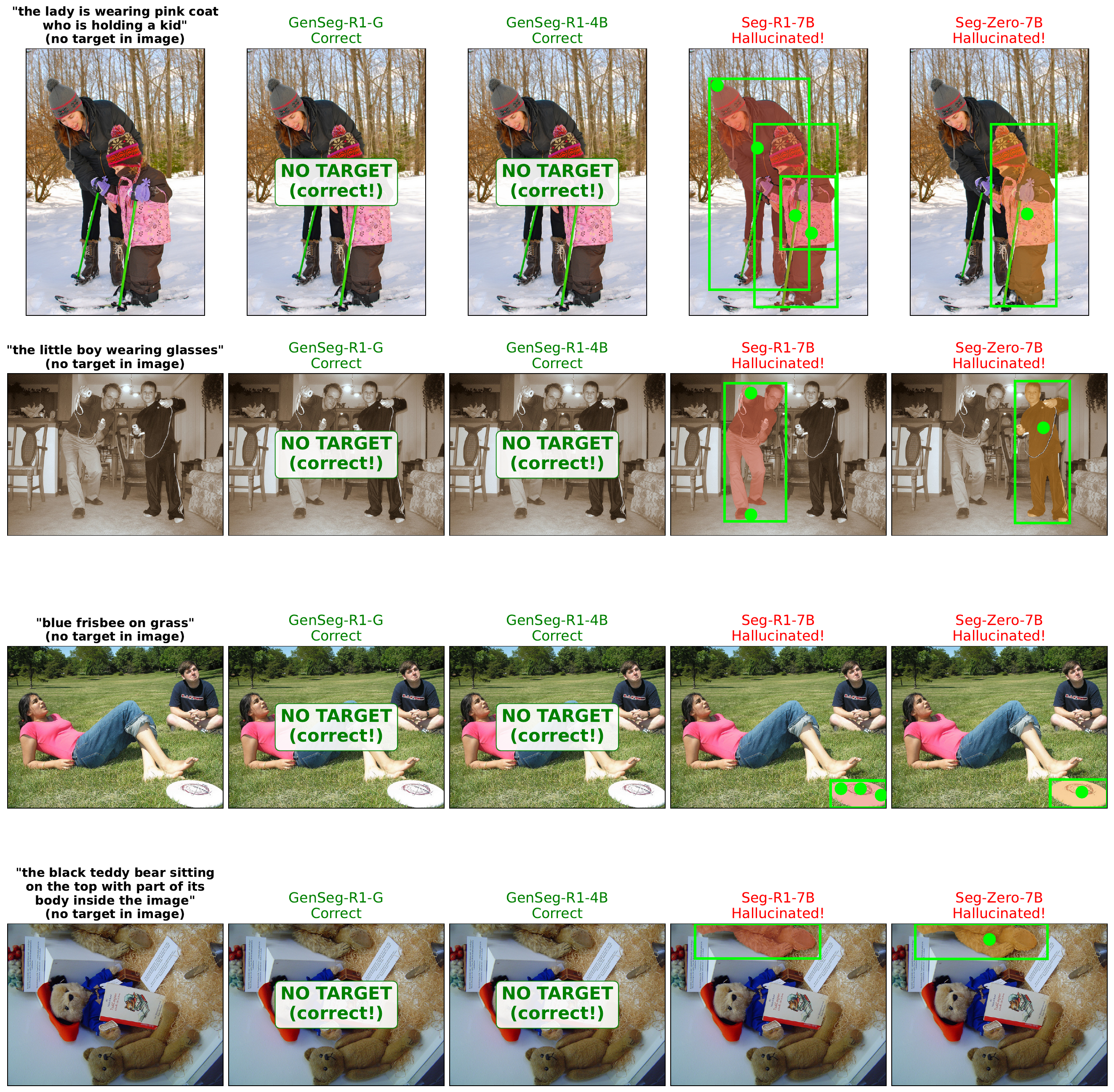}
  \caption{\textbf{No-target detection comparison.}
    For queries with no matching object, \modelname{}-G and \modelname{}-4B
    correctly predict \texttt{no\_target}, while Seg-R1-7B and Seg-Zero-7B
    hallucinate masks for non-existent objects.}
  \label{fig:qual_notarget}
\end{figure*}

\subsection{ReasonSeg Test Results}

\begin{table}[t]
  \centering
  \caption{ReasonSeg test results (779 samples).}
  \label{tab:reasonseg}
  \small
  \setlength{\tabcolsep}{10pt}
  \begin{tabular}{l cc}
    \toprule
    \textbf{Model}
      & \textbf{mIoU (\%)}~$\uparrow$
      & \textbf{cIoU (\%)}~$\uparrow$ \\
    \midrule
    \modelname{}-4B (ours)
      & \best{68.40}   & \best{62.73} \\
    \modelname{}-G\; (ours)
      & \second{65.93}  & \second{57.83} \\
    \midrule
    Seg-Zero-7B~\cite{segzero}
      & 61.41  & 55.63 \\
    Seg-R1-7B~\cite{segr1}
      & 57.75  & 48.93 \\
    \bottomrule
  \end{tabular}
\end{table}

\begin{figure*}[t]
  \centering
  \includegraphics[width=\textwidth]{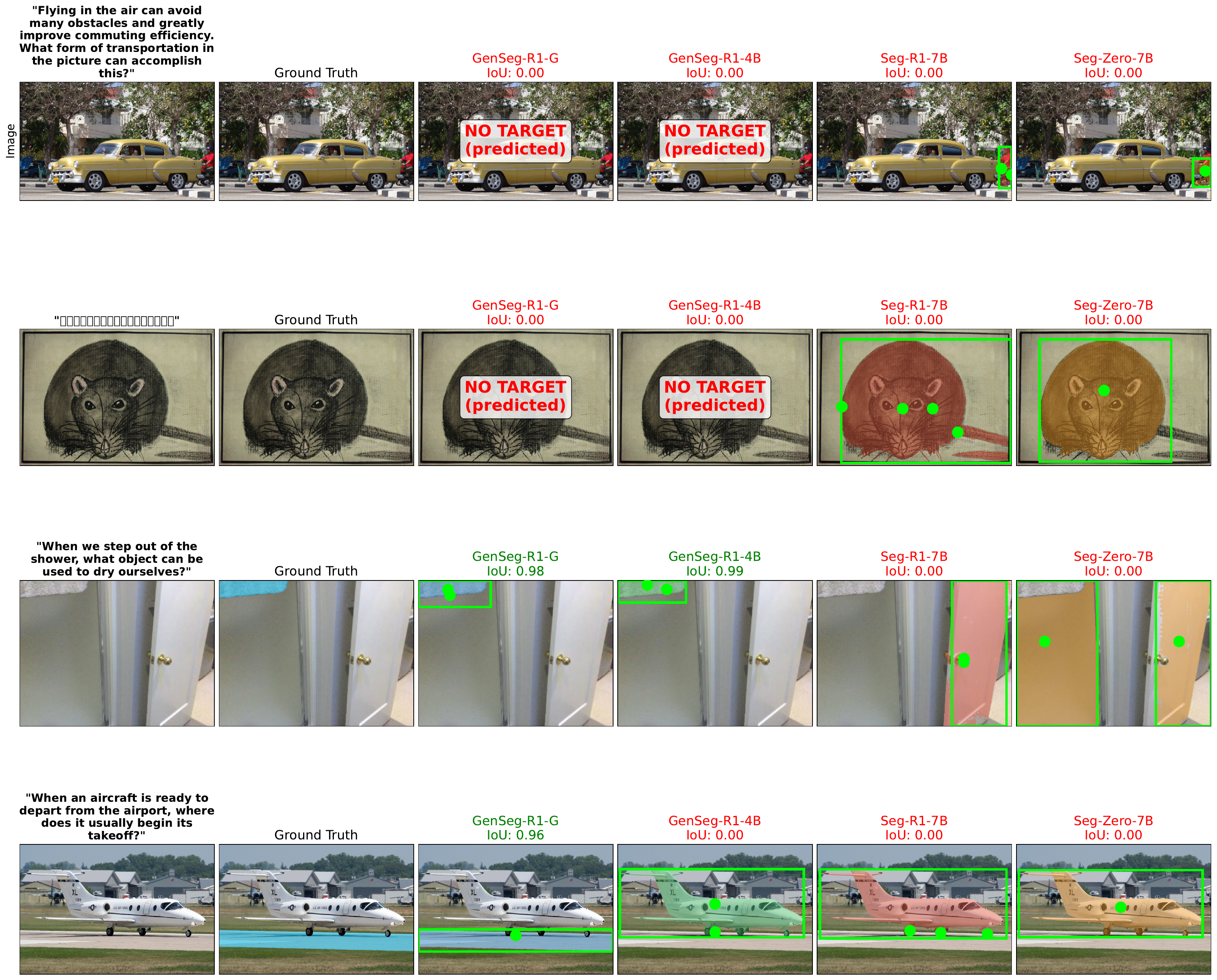}
  \caption{\textbf{Qualitative ReasonSeg comparison.}
    Each row shows a reasoning-heavy query from the ReasonSeg test set.
    The first three rows illustrate successful cases where
    \modelname{}-G and \modelname{}-4B correctly reason about implicit
    queries, while baselines struggle.  The last two rows show failure
    cases where all models fail to identify the target object.}
  \label{fig:qual_reasonseg}
\end{figure*}

ReasonSeg requires compositional and implicit reasoning (e.g.,
\emph{``the object that can be used to cut paper''}).
Table~\ref{tab:reasonseg} shows that \modelname{}-4B reaches
68.40\% mIoU, outperforming Seg-Zero-7B by \textbf{+7.0} points and
Seg-R1-7B by \textbf{+10.7} points, confirming that the emergent
\texttt{<think>} traces support non-trivial reasoning.  The
GRefCOCO-trained \modelname{}-G variant also surpasses both baselines
despite being trained on a different data distribution, achieving
65.93\% mIoU.  Figure~\ref{fig:qual_reasonseg} shows qualitative
examples of both successes and failures on reasoning-heavy queries.

\subsection{Ablation: Gain Attribution}\label{sec:ablation_gain}

Table~\ref{tab:delta} summarizes the absolute improvement brought by
GRPO finetuning for each model variant.

\begin{table}[t]
  \centering
  \caption{Absolute improvement ($\Delta$) from GRPO finetuning on
    RefCOCOg validation (mask IoU).}
  \label{tab:delta}
  \small
  \setlength{\tabcolsep}{8pt}
  \begin{tabular}{l cc cc}
    \toprule
    & \multicolumn{2}{c}{\textbf{cIoU}}
    & \multicolumn{2}{c}{\textbf{mIoU}} \\
    \cmidrule(lr){2-3}\cmidrule(lr){4-5}
    \textbf{Model pair} & Base & Ours ($\Delta$) & Base & Ours ($\Delta$) \\
    \midrule
    8B Instruct $\to$ \modelname{}-8B
      & 0.5596 & 0.7127\;{\color{samgreen}(+0.153)}
      & 0.5188 & 0.7382\;{\color{samgreen}(+0.219)} \\
    4B Instruct $\to$ \modelname{}-4B
      & 0.5019 & 0.7062\;{\color{samgreen}(+0.204)}
      & 0.4890 & 0.7318\;{\color{samgreen}(+0.243)} \\
    \bottomrule
  \end{tabular}
\end{table}

\subsection{Data Overlap Analysis}\label{sec:overlap}

\modelname{}-4B and -8B are trained on VisionReasoner-MultiObjects-7K,
which derives from COCO-family images.  \modelname{}-G is trained on
GRefCOCO, which shares underlying COCO images with RefCOCOg.  We
quantify this overlap and verify that it does not inflate RefCOCOg
results.

GRefCOCO train contains 16\,994 unique images, of which
\textbf{634} (48.8\% of RefCOCOg val's 1\,300 images) overlap.
However, only 10 out of 2\,573 RefCOCOg val referring expressions
(0.4\%) appear verbatim in GRefCOCO train---the referring expressions
themselves are almost entirely distinct.

To verify the overlap does not inflate scores, we evaluate on both the
\emph{clean} (non-overlapping, 1\,067 refs) and \emph{overlap}
(1\,506 refs) subsets separately:

\begin{table}[h]
  \centering
  \caption{Data overlap analysis: RefCOCOg cIoU on clean
    (non-overlapping) vs.\ overlap subsets.  Clean-subset cIoU is
    equal or higher, confirming no score inflation.}
  \label{tab:overlap}
  \small
  \setlength{\tabcolsep}{8pt}
  \begin{tabular}{l ccc}
    \toprule
    \textbf{Model}
      & \textbf{Full Val} (2573)
      & \textbf{Clean} (1067)
      & \textbf{Overlap} (1506) \\
    \midrule
    \modelname{}-4B cIoU & 0.7062 & \textbf{0.7069} & 0.6959 \\
    \modelname{}-8B cIoU & 0.7127 & \textbf{0.7200} & 0.6927 \\
    \bottomrule
  \end{tabular}
\end{table}

\noindent
cIoU on the clean subset is equal or \emph{higher} than on the full
val set for both models.  The \modelname{}-4B advantage over
Seg-R1-7B holds on the clean subset: \textbf{0.707 vs.\ 0.661}
(+4.6 cIoU), confirming that our gains are not attributable to data
leakage.

We note that Seg-R1~\cite{segr1} trains on DIS5K and COD10K, which
share no images with COCO, making their RefCOCOg evaluation fully
zero-shot with respect to training images.

\section{Discussion}\label{sec:discussion}

\paragraph{Why SAM\,2-in-the-loop matters.}
Training with the actual downstream segmenter in the reward loop
ensures that the VLM learns prompt formats that SAM\,2 can
effectively convert into masks.  A model optimized purely on box IoU
may produce boxes that are metrically close but positionally
sub-optimal for SAM\,2's decoder, leading to a train--test mismatch
that our approach avoids.  The \modelname{}-G variant demonstrates
this directly: by training with SAM\,2-in-the-loop reward, it
achieves 76.69\% target mIoU on GRefCOCO, substantially
exceeding the baselines.

\paragraph{Distance-based vs.\ outcome-based reward.}
\modelname{}-4B/8B use a lightweight distance-based reward (no SAM
inference during training), while \modelname{}-G uses the full
SAM\,2-in-the-loop reward.  Both are effective---the distance-based
reward achieves strong RefCOCOg results with faster training, while
the outcome-based reward additionally enables negative-prompt handling
and boundary refinement.  This suggests a practical two-stage recipe:
pre-train with the fast reward, then specialize with SAM-in-the-loop
for tasks requiring mask-level precision or no-target robustness.

\paragraph{Why negative prompts matter.}
In interactive settings users frequently issue underspecified or
incorrect queries.  A model that returns a confident mask for a
non-existent target can be worse than returning nothing.  Our
explicit no-target reward and GRefCOCO training address
this gap: \modelname{}-G achieves 82.40\% no-target accuracy while
maintaining a false-negative rate of only 0.5\%, compared to
Seg-R1 and Seg-Zero which hallucinate masks on 100\% of negative
queries.

\paragraph{Emergent reasoning.}
Despite receiving no supervised chain-of-thought data, both the 4B
and 8B models produce coherent \texttt{<think>} traces in
$>$95\% of outputs.  These traces typically describe spatial
relationships, attribute matching, and disambiguation---capabilities
that are reinforced by GRPO through the thinking bonus but not
directly supervised.

\paragraph{Limitations.}
\begin{itemize}[leftmargin=*,itemsep=2pt]
  \item \textbf{Domain shift:}\; The current training set is
        drawn from COCO-family images; performance on out-of-domain
        scenes (e.g., medical, satellite, industrial) has not been
        characterized.
  \item \textbf{Long-tail attributes:}\; Queries involving rare
        attribute compositions (``translucent green vase behind the
        curtain'') may still cause errors.
  \item \textbf{SAM\,2 sensitivity:}\; Mask quality depends on
        the prompt format; minor coordinate noise can shift SAM\,2's
        output discontinuously.
  \item \textbf{Data overlap:}\; While we demonstrate no score
        inflation from the GRefCOCO--RefCOCOg image overlap
        (Section~\ref{sec:overlap}), training on fully disjoint data
        would strengthen the zero-shot claim.
\end{itemize}

\section{Conclusion}\label{sec:conclusion}

We introduced \modelname{}, a \emph{reason-then-segment} framework
that pairs a GRPO-finetuned Qwen3-VL model with a frozen SAM\,2
segmenter.  On RefCOCOg validation, \modelname{}-8B achieves
\textbf{0.7127} cIoU and \textbf{0.7382} mIoU, outperforming
Seg-Zero-7B (+3.3 cIoU) and Seg-R1-7B (+5.2 cIoU) under identical
evaluation.

The \modelname{}-G variant extends the framework with a
SAM\,2-in-the-loop reward and GRefCOCO training, achieving
\textbf{76.69\%} target mIoU on GRefCOCO validation and
\textbf{82.40\%} no-target accuracy, while maintaining a
false-negative rate of only 0.5\%.  On ReasonSeg test,
\modelname{}-4B reaches \textbf{68.40\%} mIoU, surpassing both
Seg-Zero-7B (+7.0) and Seg-R1-7B (+10.7) by large margins.  Data
overlap analysis confirms that the RefCOCOg results are not inflated
by shared COCO images.

Future work will extend evaluation to additional reasoning-heavy
datasets, explore the two-stage reward recipe (distance pre-training
followed by SAM-in-the-loop specialization), and scale to larger
model sizes.


\end{document}